\documentclass[10pt,twocolumn,letterpaper]{article}

\usepackage[pagenumbers]{cvpr} 

\usepackage{graphicx}
\usepackage{amsmath}
\usepackage{amssymb}
\usepackage{booktabs}

\usepackage{bbding}
\usepackage{soul}
\usepackage{bibunits}
\usepackage{bm}
\usepackage{array}
\usepackage{xcolor}
\usepackage{pdflscape}
\usepackage{makecell}
\usepackage{framed,multirow}
\usepackage[flushleft]{threeparttable}
\usepackage{amssymb}
\usepackage{pifont}
\usepackage{algorithm}
\usepackage{listings}
\usepackage{pythonhighlight}
\makeatletter
\newcommand\footnoteref[1]{\protected@xdef\@thefnmark{\ref{#1}}\@footnotemark}
\makeatother

\newcolumntype{P}[1]{>{\centering\arraybackslash}p{#1}}
\newlength\savewidth\newcommand\shline{\noalign{\global\savewidth\arrayrulewidth
  \global\arrayrulewidth 0.8pt}\hline\noalign{\global\arrayrulewidth\savewidth}}

\newcommand{\authorskip}{\hspace{2.5mm}}

\definecolor{Highlight}{HTML}{39b54a}  

\newcolumntype{x}[1]{>{\centering\arraybackslash}p{#1pt}}
\newcolumntype{z}[1]{>{\raggedright\arraybackslash}p{#1pt}}

\newcommand{\methodname}{{\fontfamily{ppl}\selectfont
SQUID}}
\newcommand{\ourdataset}{{\fontfamily{ppl}\selectfont
DigitAnatomy}}

\usepackage[pagebackref,breaklinks,colorlinks]{hyperref}

\usepackage[capitalize]{cleveref}
\crefname{section}{Sec.}{Secs.}
\Crefname{section}{Section}{Sections}
\Crefname{table}{Table}{Tables}
\crefname{table}{Tab.}{Tabs.}

\def\cvprPaperID{2427} 
\def\confName{CVPR}
\def\confYear{2023}

\begin{document}

\title{Deep Feature In-painting for Unsupervised Anomaly Detection in Radiography Images}
\title{Deep Feature In-painting for Unsupervised Anomaly Detection in X-ray Images}
\title{SQUID: Deep Feature In-Painting for Unsupervised Anomaly Detection}

\author{
 Tiange Xiang\textsuperscript{1} \authorskip Yixiao Zhang\textsuperscript{2} \authorskip Yongyi Lu\textsuperscript{2} \authorskip Alan L. Yuille\textsuperscript{2}\\Chaoyi Zhang\textsuperscript{1} \authorskip Weidong Cai\textsuperscript{1} \authorskip Zongwei Zhou\textsuperscript{2,}\thanks{Corresponding author: Zongwei Zhou (\href{mailto:zzhou82@jh.edu}{zzhou82@jh.edu})} \\[2.5mm]
 \textsuperscript{1}University of Sydney \quad \textsuperscript{2}Johns Hopkins University\\[1.5mm]
 {\small GitHub:~\href{https://github.com/tiangexiang/SQUID}{https://github.com/tiangexiang/SQUID}}
}

\maketitle

\begin{abstract}

Radiography imaging protocols focus on particular body regions, therefore producing images of great similarity and yielding recurrent anatomical structures across patients.
To exploit this structured information, we propose the use of \ul{S}pace-aware Memory \ul{Qu}eues for \ul{I}n-painting and \ul{D}etecting anomalies from radiography images (abbreviated as \methodname).
We show that \methodname\ can taxonomize the ingrained anatomical structures into recurrent patterns; and in the inference, it can identify anomalies (unseen/modified patterns) in the image.
\methodname\ surpasses 13 state-of-the-art methods in unsupervised anomaly detection by at least 5 points on two chest X-ray benchmark datasets measured by the Area Under the Curve (AUC).
Additionally, we have created a new dataset (\ourdataset), which synthesizes the spatial correlation and consistent shape in chest anatomy. We hope \ourdataset\ can prompt the development, evaluation, and interpretability of anomaly detection methods.



\end{abstract}

\section{Introduction}
\label{sec:intro}

\begin{figure}[t]
    \centering
    \includegraphics[width=1.0\linewidth]{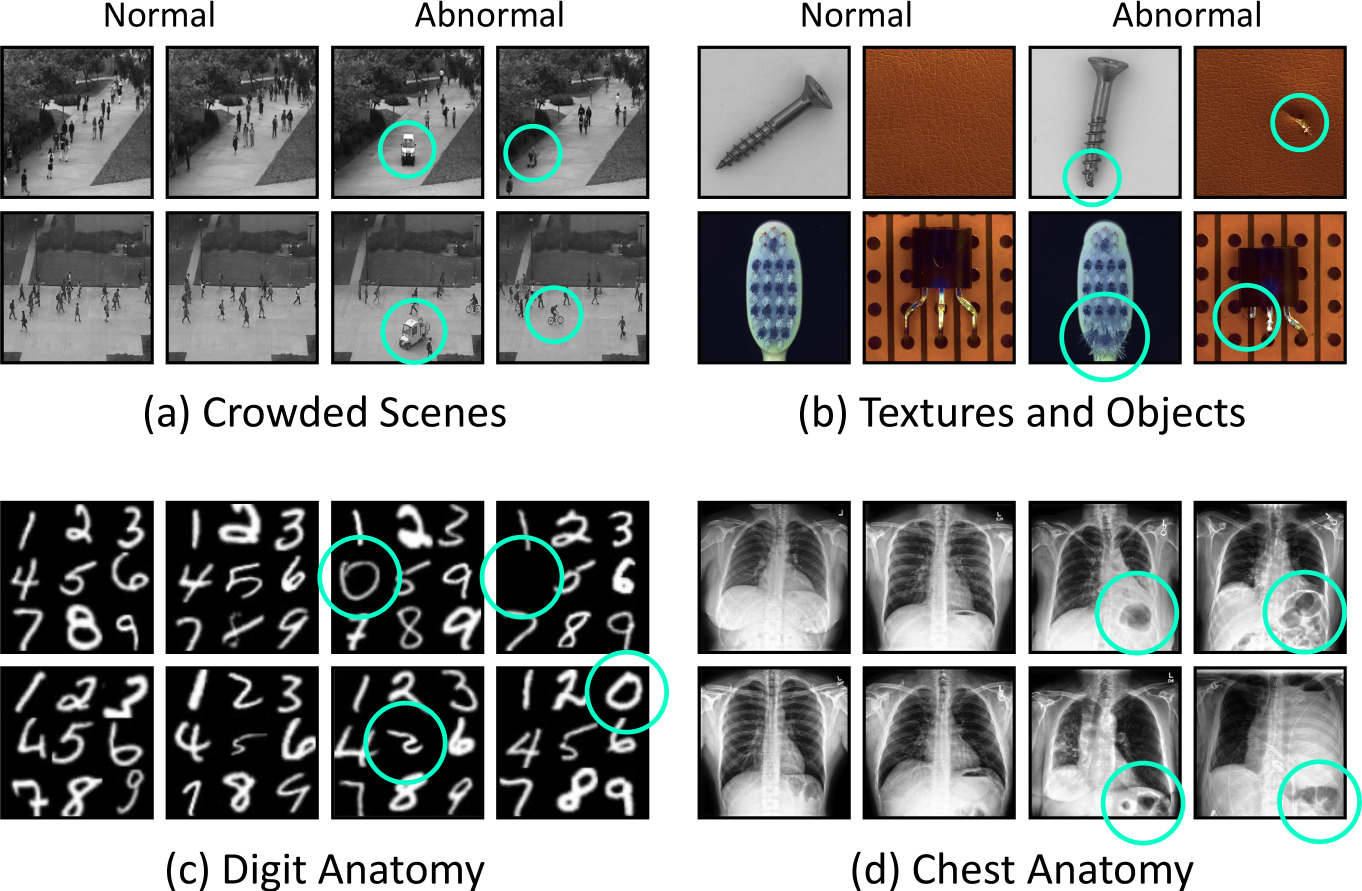}
    \caption{
    Anomaly detection in  radiography images can be both easier and harder than in photographic images.
    It is easier because  radiography images are spatially structured due to consistent imaging protocols.
    It is harder because anomalies in  radiography images are subtle and require medical expertise to annotate.
    }
    \label{fig:introductory}
\end{figure}

Vision tasks in photographic imaging and radiography imaging are different. For example, when identifying objects in photographic images, we assume translation invariance---a cat is a cat no matter if it appears on the left or right of the image. 
In radiography imaging, on the other hand, the relative location and orientation of a structure are important characteristics that allow the identification of normal anatomy and pathological conditions~\cite{zhao20213d,haghighi2021transferable}. 
Since radiography imaging protocols assess patients in a fairly consistent orientation, the generated images have great similarity across various patients, equipment manufacturers, and facility locations (see examples in \figureautorefname~\ref{fig:introductory}d). 
The consistent and recurrent anatomy facilitates the analysis of numerous critical problems and should be considered a significant advantage for radiography imaging~\cite{zhou2021towards}.
Several investigations have demonstrated the value of this prior knowledge in enhancing Deep Nets' performance by adding location features, modifying objective functions, and constraining coordinates relative to landmarks in images~\cite{smoger2015statistical,anas2016automatic,mirikharaji2018star,zhou2019integrating,lu2020learning}. 
Our work seeks to answer this critical question: 
\emph{Can we exploit consistent anatomical patterns and their spatial information to strengthen Deep Nets' detection of anomalies from  radiography images without manual annotation?}

Unsupervised anomaly detection only uses healthy images for model training and requires no other annotations such as disease diagnosis or localization~\cite{baur2021autoencoders}.
As many as 80\% of clinical errors occur when the radiologist misses the abnormality in the first place \cite{brady2017error}. The impact of anomaly detection is to reduce that 80\% by clearly pointing out to radiologists that there exists a suspicious lesion and then having them look at the scan in depth. Unlike previous anomaly detection methods, we formulate the task as an in-painting task to exploit the anatomical consistency in appearance, position, and layout across radiography images. 
Specifically, we propose \ul{S}pace-aware Memory \ul{Qu}eues for \ul{I}n-painting and \ul{D}etecting anomalies from radiography images (abbreviated as \methodname).
During training, our model can \emph{dynamically} maintain a visual pattern dictionary by taxonomizing recurrent anatomical patterns based on their spatial locations. 
Due to the consistency in anatomy, the same body region across healthy images is expected to express similar visual patterns, which makes the total number of unique patterns manageable.
During inference, since anomaly patterns do not exist in the dictionary, the generated  radiography image is expected to be unrealistic if an anomaly is present.
As a result, the model can identify the anomaly by discriminating the quality of the in-painting task.
The success of anomaly detection has two basic assumptions~\cite{zimek2017outlier}: \emph{first}, anomalies only occur rarely in the data;
\emph{second}, anomalies differ from the normal patterns significantly.

We have conducted experiments on two large-scale, publicly available radiography imaging datasets. 
Our \methodname\ is significantly superior to predominant methods in unsupervised anomaly detection by over 5 points on the ZhangLab dataset~\cite{kermany2018identifying}; remarkably, we have demonstrated a 10-point improvement over 13 recent unsupervised anomaly detection methods on the Stanford CheXpert dataset~\cite{irvin2019chexpert}. In addition, we have created a new dataset (\ourdataset) to elucidate \textit{spatial correlation} and \textit{consistent shape} of the chest anatomy in radiography (see \figureautorefname~\ref{fig:introductory}c). 
\ourdataset\ is dedicated to easing the development, evaluation, and interpretability of anomaly detection methods.
The qualitative visualization clearly shows the superiority of our \methodname\ over the current state-of-the-art methods.

In summary, our contributions include: 
\textbf{(I)} the best performing unsupervised anomaly detection method for chest  radiography imaging;
\textbf{(II)} a synthetic dataset to promote anomaly detection research;
\textbf{(III)} \methodname\ overcomes limitations in dominant unsupervised anomaly detection methods~\cite{kingma2013auto,akcay2018ganomaly,schlegl2019f,gong2019memorizing,zhao2021anomaly} by inventing 
Space-aware Memory Queue (\S\ref{sec:queue}), and
Feature-level In-painting (\S\ref{sec:inpaint}).

\section{Related Work}
\label{sec:related_work}

\smallskip\noindent\textbf{\textit{Anomaly detection in natural imaging.}}
Anomaly detection is the task of identifying rare events that deviate from the distribution of normal data~\cite{omar2013machine}. 
Early attempts include one-class SVM~\cite{scholkopf1999support}, dictionary learning~\cite{zhao2011online}, and sparse coding~\cite{cong2011sparse}.
Due to the lack of sufficient samples of anomalies, later works typically formulate anomaly detection as an unsupervised learning problem~\cite{ruff2018deep,zong2018deep,sidibe2017anomaly,hendrycks2016baseline,lee2017training,liang2017enhancing,lee2018simple,devries2018learning,hendrycks2018deep}. These can be roughly categorized into reconstruction-based and density-based methods. Reconstruction-based methods train a model (\eg Auto-Encoder) to recover the original inputs~\cite{chen2018unsupervised,siddiquee2019learning,zhou2019models,tang2021disentangled,zhou2021models,xiao2022delving} and identify anomalies by analyzing reconstruction errors. Density-based methods predict anomalies by estimating the normal data distribution (\eg via VAEs~\cite{kingma2013auto} or GANs~\cite{schlegl2017unsupervised,schlegl2019f,alex2017generative}). However, the learned distribution for normal images by these methods cannot explain the possible abnormalities. In this paper, we address these limitations by maintaining a visual pattern memory from homogeneous medical images. 
Several other previous works investigated the use of image in-painting for anomaly detection, \ie parts of the input image are masked out, and the model is
trained to recover the missing parts in a self-supervised way~\cite{reiss2021panda,nguyen2021unsupervised,zavrtanik2021reconstruction,haselmann2018anomaly,li2021cutpaste}. 
There are also plenty of works on detecting anomalies in video sequences~\cite{lu2019future,esser2018variational,liu2021hybrid}. Bergmann~\etal~\cite{bergmann2020uninformed} and Salehi~\etal~\cite{salehi2020distillation} proposed similar student-teacher networks, whereas our method utilizes such a structure to distillate input-aware features only, and the teacher network is completely disabled during inference. 

\smallskip\noindent\textbf{\textit{Anomaly detection in medical imaging.}}
Anomaly detection in the medical domain is usually approached on a per-pathology basis. Supervised learning based methods~\cite{schlegl2015predicting,baur2021autoencoders,liu2023clip} are commonly adopted to detect specific types of abnormalities, such as lesions~\cite{zuluaga2011learning}, pathologies~\cite{khan2021attributes}, tumors~\cite{bakas2018identifying}, and nodules~\cite{zheng2019automatic}. Recent unsupervised methods have been proposed to detect anomalies in general~\cite{siddiquee2019learning,baur2021autoencoders,heer2021ood}, 
With the help of GANs, anomaly detection can be achieved with \textit{weak} annotations. In AnoGAN~\cite{schlegl2017unsupervised}, the discriminator was heavily over-fitted to the normal image distribution to detect the anomaly. Subsequently, f-AnoGAN~\cite{schlegl2019f} was proposed to improve computational efficiency. Marimont~\etal~\cite{naval2021implicit} designed an auto-decoder network to fit the distribution of normal images. The spatial coordinates and anomaly probabilities are mapped over a proxy for different tissue types. Han~\etal~\cite{han2021madgan} proposed a two-step GAN-based framework for detecting anomalies in MRI slices as well. However, their method relies on a voxel-wise representation for the 3D MRI sequences, which is impossible in our task. Most recently, a hybrid framework SALAD~\cite{zhao2021anomaly} was proposed that combines GAN with self-supervised techniques. Normal images are first augmented to carry the forged anomaly through pixel corruption and pixel shuffling. The fake abnormal images, along with the original normal ones, are fed to the GAN for learning more robust feature representations. However, these approaches demand strong prior knowledge and assumptions about the anomaly type to make the augmentation effective. 
Differing from photographic images, radiography imaging protocols produce images with consistent anatomical patterns, which are much more challenging to detect due to subtle imaging clues and overlapping anatomic structures (\figureautorefname~\ref{fig:introductory}). 
Unlike most existing works, we present a novel method that explicitly harnesses the radiography images' properties, dramatically improving the performance in anomaly detection from radiography images.

\begin{figure*}[t]
    \centering
    \includegraphics[width=1.0\linewidth]{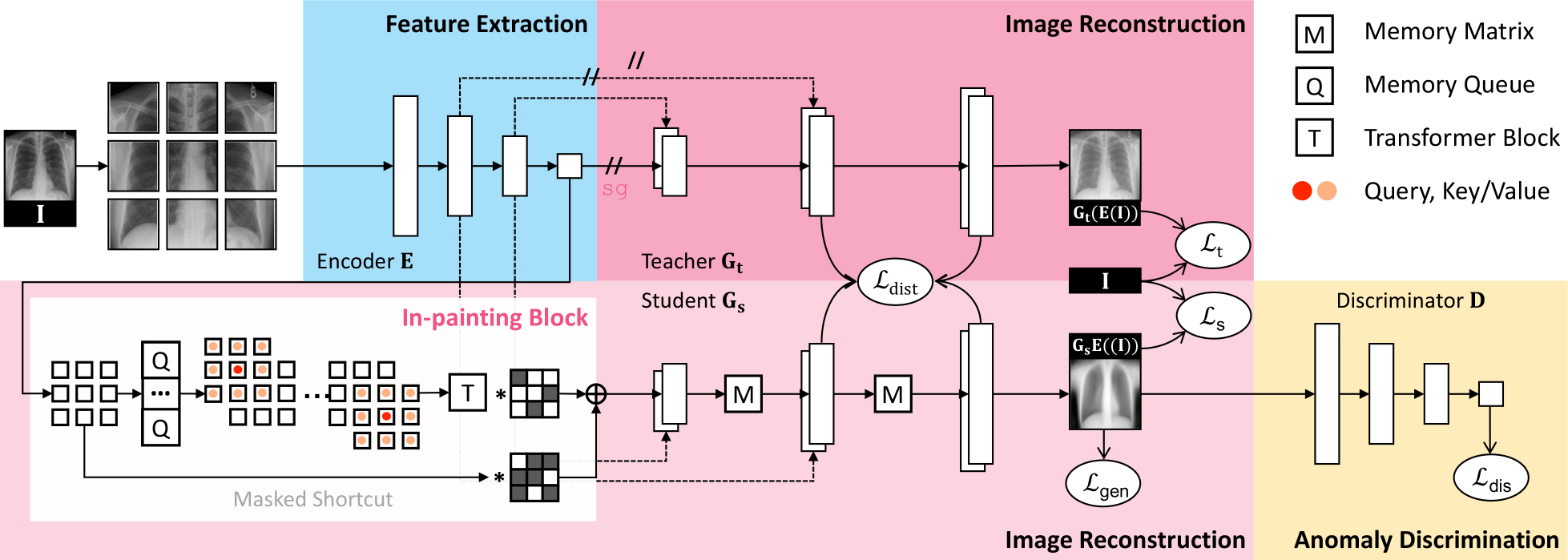}
    \caption{\textbf{\methodname.} We divide an input image into $N\times N$ non-overlapping patches and feed them into the encoder for feature extraction. 
    Two generators will be trained to reconstruct the original image. 
    Along with the reconstruction, a dictionary of anatomical patterns will be created and updated dynamically via a novel Memory Queue (\S\ref{sec:queue});
    The teacher generator directly uses the features extracted by the encoder;
    the student generator uses the features augmented by our in-painting block (\S\ref{sec:inpaint}). 
    The teacher and student generators are coupled through a knowledge distillation paradigm.
    We employ a discriminator to assess whether the image reconstructed by the student generator is real or fake.
    Once trained, it can also be used to detect anomalies in test images (\S\ref{sec:alert}).
    }
    \label{fig:framework}
\end{figure*}

\smallskip\noindent\textbf{\textit{Memory networks.}}
Incorporating memory modules into neural networks has been demonstrated to be effective for many tasks ~\cite{kumar2016ask,fan2019heterogeneous,kaiser2017learning,cai2018memory,lee2018memory}. Adopting Memory Matrix for unsupervised anomaly detection was first proposed in MemAE~\cite{gong2019memorizing}. In addition to auto-encoding (AE), an extra Memory Matrix was introduced between the encoder and the decoder to capture normal feature patterns during training. The matrix is jointly optimized along with the AE and hence learns an essential basis to be able to assemble normal patterns. Based on this paradigm, Park~\etal~\cite{park2020learning} introduced a non-learnable memory module that can be updated with inputs. Note that although our proposed Memory Queue also does not require any gradients, our method differs significantly in its usage purpose and updating rules. Considering the extra memory usage in existing methods, Lv~\etal~\cite{Lv2021MPN} proposed a dynamic prototype unit that encodes normal dynamics on the fly, while consuming little additional memory.
In this paper, we overcome the limitations of the Memory Matrix and propose an effective yet efficient Memory Queue for unsupervised anomaly detection in  radiography images.

\section{\methodname}

\subsection{Overview} \label{sec:framework}

\noindent\textbf{(1) \textit{Feature extraction.}} 
We divide the input image into $N\times N$ non-overlapping patches and feed them into an encoder for feature extraction. 
The extracted features will be used for image reconstruction. 
Practically, the encoder can be any backbone architectures~\cite{tan2019efficientnet,dosovitskiy2020image}; we adopt basic Convolutions and Pooling layers in this work for simplicity. 

\smallskip\noindent\textbf{(2) \textit{Image reconstruction.}} 
We introduce teacher and student generators to reconstruct the original image. 
Along with the reconstruction, a dictionary of anatomical patterns will be created and updated dynamically as a \textbf{Memory Queue} (\S\ref{sec:queue}).
Specifically, the teacher generator directly reconstructs the image using the features extracted by the encoder (essentially an auto-encoder~\cite{rumelhart1985learning}). 
The student generator, on the other hand, using the features augmented by our \textbf{in-painting block} (\S\ref{sec:inpaint}). 
The teacher and student generators are coupled through a knowledge distillation paradigm~\cite{hinton2015distilling} at all of the up-sampling levels.
The objective of the student generator is to reconstruct a normal image from the augmented features, which will then be used for anomaly discrimination (\S\ref{sec:alert});
while the teacher generator\footnote{We disabled the backpropagation between the teacher and encoder by stop-gradient~\cite{he2020momentum} and showed its empirical benefit in \tableautorefname~\ref{tab:component}.} serves as a regularizer that prevents the student from constantly generating the same normal image.

\smallskip\noindent\textbf{(3) \textit{Anomaly discrimination.}}
Following the adversarial learning~\cite{schlegl2017unsupervised,schlegl2019f}, we employ a discriminator to assess whether the generated image is real or fake.
Only the student generator will receive the gradient derived from the discriminator.
The two generators and the discriminator are competing against each other until they converge to an equilibrium.
Once trained, the discriminator can be used to detect anomalies in test images (\S\ref{sec:alert}).

\begin{figure}[t]
    \centering
    \includegraphics[width=1.0\linewidth]{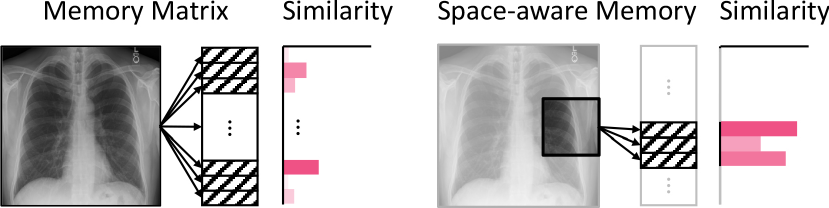}
    \caption{\textbf{Space-aware memory.} For unique encoding of location information, we restrict each patch to be only accessible by a non-overlapping region in the memory.}
    \label{fig:space_mem}
\end{figure}

\subsection{Inventing Memory Queue as Dictionary}
\label{sec:queue}

\noindent\textbf{\textit{Motivation.}} 
The Memory Matrix was introduced by Gong \etal \cite{gong2019memorizing} and has since been widely adopted in unsupervised anomaly detection~\cite{liu2021hybrid,zaheer2021anomaly,gong2021memory}. 
To forge a ``normal'' appearance, features are \textit{augmented} by weighted averaging the similar patterns in Memory Matrix.
This augmentation is, however, applied to the features extracted from the whole image, discarding the spatial information in images.
Therefore, the Memory Matrix in its current form cannot perceive the anatomical consistency as in  radiography images.

\smallskip\noindent\textbf{\textit{Space-aware memory.}} 
To harness the spatial information, we pass the divided small patches, instead of the whole image, into the model.
These patches are associated with unique location identifiers of the original image.
We seek to build the relationship between the patch location and memory region, by restricting the search space in Memory Matrix to the patch-corresponded non-overlapping segments only.
That is, a patch at a particular location can only access a corresponding segment in the whole Memory Matrix (illustrated in~\figureautorefname~\ref{fig:space_mem}).
We refer to this new strategy as ``space-aware memory'' because it enables explicit encoding of the spatial information into Memory Matrix.
Space-aware memory can also accelerate the augmentation speed compared with~\cite{gong2019memorizing} as it no longer goes through the entire Memory Matrix to assemble similar features.

\begin{figure}[t]
    \centering
    \includegraphics[width=1.0\linewidth]{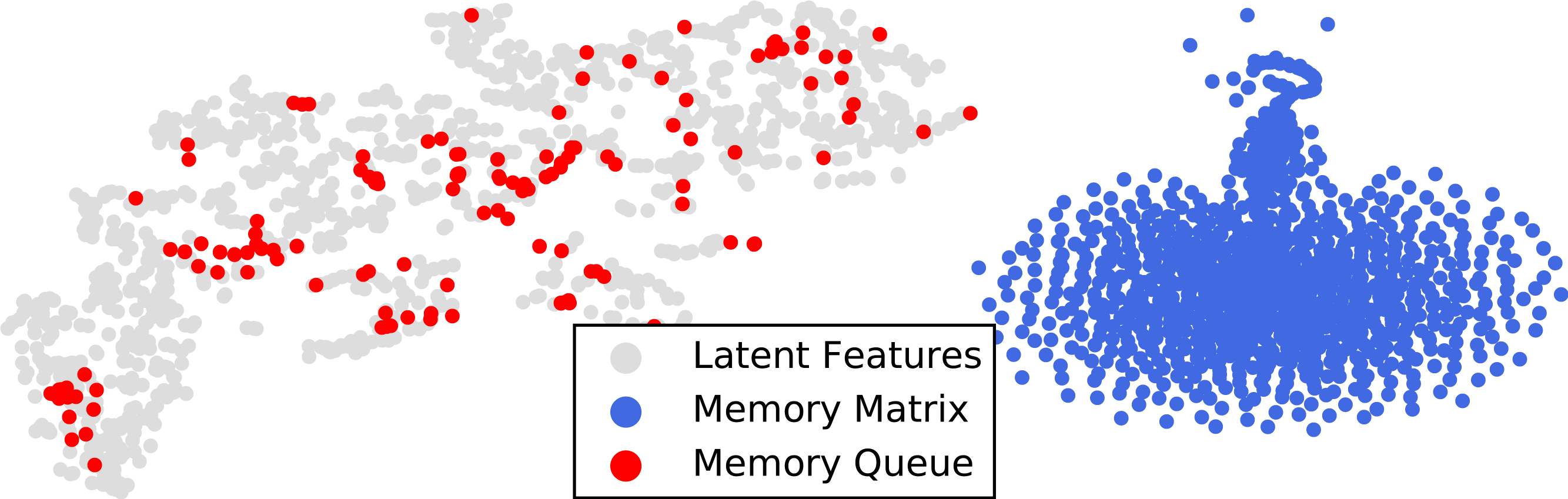}
    \caption{t-SNE visualizations of patterns in Memory Matrix, Memory Queue, and patch features of the training samples~\cite{van2008visualizing}. Patterns in the Memory Matrix are far away from the distribution of patch features, while patterns in the Memory Queue (as copies of previously seen features) share a similar distribution.}
    \label{fig:tsne}
\end{figure}

\begin{figure*}[t]
    \centering
    \includegraphics[width=1.0\linewidth]{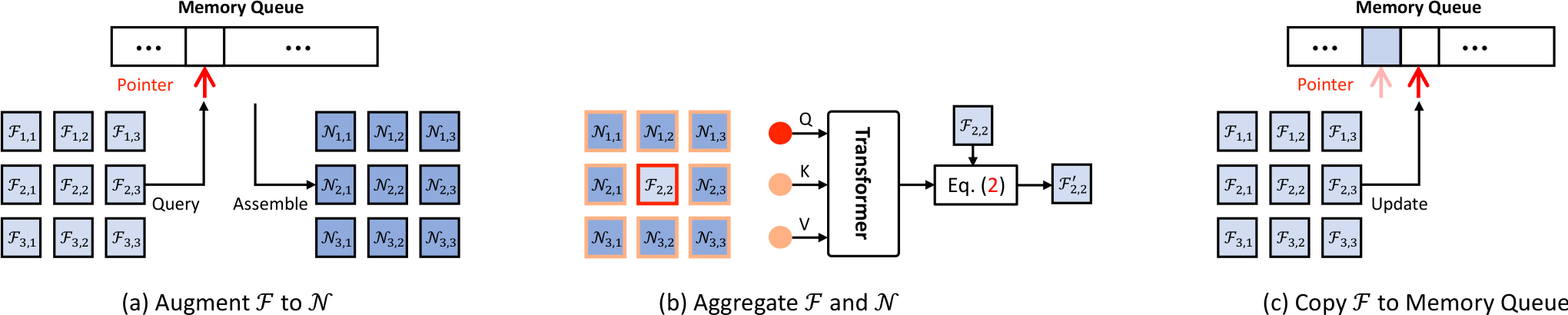}
    %
    \caption{\textbf{Three-step workflow of our in-painting block.} (a) Each non-overlapping patch feature $\mathcal{F}$ is queried to a unique region in Memory Queue, and the most similar items are assembled to $\mathcal{N}$. (b) Each center patch feature $\mathcal{F}$ and its eight neighbors $\mathcal{N}$ are used as query and key/value, respectively, to a Transformer layer for in-painting. (c) Each Memory Queue region copies its corresponding patch features $\mathcal{F}$ into the memory by maintaining a pointer. Note that this step is only performed during training.}
    \label{fig:inpaint_block}
\end{figure*}

\smallskip\noindent\textbf{\textit{Memory queue.}} 
In learning-based Memory Matrix~\cite{gong2019memorizing}, ``normal patterns'' are forged by combining learned basis in the matrix. However, there is always a distribution discrepancy between the basic combinations and the actual image features. This disparity makes it hard for the subsequent image generation.
To address this issue, we propose a Memory Queue to store \emph{real} image features during model training, therefore presenting an identical distribution to the image features.
Specifically, it directly copies previously seen features into a queue structure during training\footnote{In practice, copying features into the queue at every training iteration demands considerable computational time. 
Supposing $N$ patterns in the queue and $M$ training iterations, the sampling strategy in~\cite{gong2019memorizing} demands an $\mathcal{O}(NM)$ time complexity. 
We implement it more efficiently: at each iteration, the current batch of features will be copied into the queue for \emph{only once} (\figureautorefname~\ref{fig:inpaint_block}c), yielding a linear complexity of $\mathcal{O}(M + cM)$ with the copy-and-paste operation in a constant time $c$.
We follow the first-in-first-out (FIFO) paradigm to update the queue continuously.}. 
Once trained, Memory Queue can be used as a \textit{dictionary} of normal anatomical patterns. 
In \figureautorefname~\ref{fig:tsne}, we show t-SNE visualizations to validate that the learned basis in Memory Matrix (blue dots) distributes differently from the actual image features of the training set (gray dots). In contrast, the stored image features in our Memory Queue (red dots) are in an identical distribution to the actual ones.

\smallskip\noindent\textbf{\textit{Gumbel shrinkage.}}
Controlling the number of activated patterns in the memory has proven to be advantageous for anomaly detection~\cite{graves2014neural,gong2019memorizing}.
However, setting a hard shrinkage threshold fails to adapt to cases where no suitable entries can be found in the memory. One natural workaround is to activate the top-$k$ similar patterns in the memory. However, this strategy restricts the gradient flow to only the top-$k$ memory entries, while the rest inactivated ones could not receive any gradients and be updated as expected.
To extend gradients to all patterns in the memory, inspired by Jang~\cite{jang2016categorical}, we present a \textit{Gumbel Shrinkage} schema:
\begin{equation} 
\label{eq:gumbel_shrinkage}
    \begin{split}
        \mathbf{w'}  = & \ \texttt{sg}(\texttt{hs}(\mathbf{w}, \texttt{topk}(\mathbf{w})) - \phi(\mathbf{w})) + \phi(\mathbf{w}),
\end{split}
\end{equation}
where $\mathbf{w}$ denotes the similarity between the image features and entries in Memory, $\texttt{sg}(\cdot)$ the stop-gradient operation, $\texttt{hs}(\cdot,t)$ the hard shrinkage operator with threshold $t$, and $\phi(\cdot)$ the Softmax function. In the forward pass, Gumbel Shrinkage ensures the combination of the top-$k$ most similar entries in the memory;
During the back-propagation, Gumbel Shrinkage essentially functions as Softmax. 
We apply Gumbel Shrinkage to both Memory Queue and Memory Matrix in our framework.

\subsection{Formulating Anomaly Detection as In-painting}
\label{sec:inpaint}

\noindent\textbf{\textit{Motivation.}}
Image in-painting~\cite{li2020recurrent,pathak2016context} was initially proposed to recover corrupted image regions with neighboring context.
Following the above intuition, we propose to achieve anomaly detection via in-painting anomalous radiography patterns into healthy ones.
When in-painting pixels in the image space, recovered regions have been usually seen to associate with boundary artifacts, particularly when using Deep Nets~\cite{liu2018image}.
These undesired artifacts are responsible for numerous false positives when formulating anomaly detection as a pixel-level in-painting task~\cite{siddiquee2019learning,zhou2021models}. 
To alleviate this issue, we achieve the in-painting task at the feature level instead. Latent features are better invariant to pixel-level noise, rotation, and translation, therefore are more suitable for subsequent anomaly detection.

\smallskip\noindent\textbf{\textit{In-painting block.}} 
We integrate our Memory Queue inside a novel in-painting block to perform feature-space in-painting. The block starts with a Memory Queue that augments $w\times h$ non-overlapping patch features $\mathcal{F}_{\{(1,1),\cdots,(w,h)\}}$ into their most similar ``normal'' patterns $\mathcal{N}_{\{(1,1),\cdots,(w,h)\}}$ (\figureautorefname~\ref{fig:inpaint_block}a). 
Since $\mathcal{N}$ is assembled by features extracted from previous training data, $\mathcal{N}$ is not subject to the current input image. 
To recap the characteristics of the input image, we aggregate both patch features $\mathcal{F}$ and their augmented features $\mathcal{N}$ using a transformer block~\cite{vaswani2017attention}. In details, for each patch $\mathcal{F}_{i,j}$, its spatially adjacent eight augmented ones $\mathcal{N}_{\{(i-1,j-1),\cdots,(i+1,j+1)\}}$ are used as conditions to refine $\mathcal{F}_{i,j}$ (\figureautorefname~\ref{fig:inpaint_block}b).
The query token of the transformer block is flattened $\mathcal{F}_{(i,j)}\in\mathcal{R}^{1\times *}$ and key/value tokens are $\mathcal{N}_{\{(i-1,j-1),\cdots,(i+1,j+1)\}}$ $\in \mathcal{R}^{8\times *}$. 
At the start and the end of our in-painting block, we apply an extra pair of point-wise convolutions ($1\times 1$ convolutional kernel)~\cite{he2016deep}.

\smallskip\noindent\textbf{\textit{Masked shortcut.}} 
We employ a shortcut within the in-painting block to better aggregate features and ease optimization. 
Our empirical study shows that a direct residual connection downgrades the effectiveness of the in-painting block (Appendix~\ref{sec:extensive_ablation_appendix}). 
Inspired by Xiang~\etal~\cite{xiang2021partial}, we utilize a random binary mask to gate shortcut features during training (\figureautorefname~\ref{fig:inpaint_block}b). As such, given the input patch features $\mathcal{F}$, the output of the in-painting block is obtained by:
\begin{equation} \label{eq:masked_shortcut}
	\mathcal{F}' = (1-\delta)\cdot \mathcal{F} + \delta\cdot\texttt{inpaint}(\mathcal{F}),
\end{equation} 
where $\texttt{inpaint}(\cdot)$ is the designed in-painting block, $\delta\sim\emph{Bernoulli}(\rho)$ is a binary variable with $\rho$ the gating probability. After obtaining $\mathcal{F}'$ at each training step, the originally $\mathcal{F}$ are then copied to update the memory (\figureautorefname~\ref{fig:inpaint_block}c). During inference, we disable the shortcut completely such that $\mathcal{F}'=\texttt{inpaint}(\mathcal{F})$ for deterministic predictions.

\subsection{Anomaly Discrimination} \label{sec:alert}

Our discriminator can detect anomalies by assessing the quality of the reconstructions---normal if realistic; abnormal otherwise. It is because the generator was trained on normal images, so Memory Queues only store normal patterns. During inference, since abnormal patterns were never present in Memory Queues, the reconstructed image is expected to appear unrealistic.

Our in-painting block focuses on augmenting any patch feature (either normal or abnormal) into similar ``normal'' features. 
The student generator then reconstructs a ``normal'' image based on the ``normal'' features.
The teacher generator is used to prevent the student from generating the same image regardless of inputs.
Once trained, the semantic (rather than pixel-level) difference between the input and the student generator's reconstructed image is expected to be small if normal and big otherwise.
We, therefore, delegate the optimized discriminator network for alerting anomalies perceptually. For better clarification, we notate the encoder, teacher generator, student generator, and discriminator as $\mathbf{E}$, $\mathbf{G}_{\text{t}}$, $\mathbf{G}_{\text{s}}$, and $\mathbf{D}$. An anomaly score $A$ can be computed through:
$A = \phi(\frac{\mathbf{D}(\mathbf{G}_{\text{s}}(\mathbf{E}(\mathbf{I}))) - \mu}{\sigma})$,
where $\phi(\cdot)$ is the Sigmoid function, $\mu$ and $\sigma$ are the mean and standard deviation of anomaly scores calculated on training samples. 


\subsection{Loss Function} \label{sec:loss}

\methodname\ is optimized by five loss functions. 
The mean square error (MSE) between input and reconstructed images is used for both teacher and student generators.
Concretely, $\mathcal{L}_{\text{t}} = (\mathbf{I} - \mathbf{G}_{\text{t}}(\mathbf{E}(\mathbf{I})))^2$ and  $\mathcal{L}_{\text{s}} = (\mathbf{I} - \mathbf{G}_{\text{s}}(\mathbf{E}(\mathbf{I})))^2$ for the teacher and student generators, respectively, where $\mathbf{I}$ denotes the input image.
Following the knowledge distillation paradigm, we apply a distance constraint between the teacher and student generators to all levels of features:
$\mathcal{L}_{\text{dist}} = \sum^{l}_{i=1}(\mathcal{F}_{\text{t}}^i - \mathcal{F}_\text{s}^i)^2$,
where $l$ is the level of features used for knowledge distillation, $\mathcal{F}_{\text{t}}$ and $\mathcal{F}_{\text{s}}$ are the intermediate features in the teacher and student generators, respectively.
In addition, we employ an adversarial loss (similar to DCGAN~\cite{radford2015unsupervised}) to improve the quality of the image generated by the student generator.
Specifically, the following equation is minimized: $\mathcal{L}_{\text{gen}} = \log(1-\mathbf{D}(\mathbf{G}_{\text{s}}(\mathbf{E}(\mathbf{I}))))$. 
The discriminator seeks to maximize the average of the probability for real images and the inverted probability for fake images: $\mathcal{L}_{\text{dis}} = \log(\mathbf{D}(\mathbf{I})) + \log(1-\mathbf{D}(\mathbf{G}_{\text{s}}(\mathbf{E}(\mathbf{I}))))$. In summary, \methodname\ is trained to \emph{minimize} the generative loss terms ($\lambda_{\text{t}}\mathcal{L}_{\text{t}} + \lambda_{\text{s}}\mathcal{L}_{\text{s}} + \lambda_{\text{dist}}\mathcal{L}_{\text{dist}} + \lambda_{\text{gen}}\mathcal{L}_{\text{gen}}$)
and to \emph{maximize} the discriminative loss term ($\lambda_{\text{dis}}\mathcal{L}_{\text{dis}}$).

\section{Experiments}

\subsection{New Benchmark}
\label{sec:new_benchmark}

\begin{figure}[t]
    \centering
    \includegraphics[width=1.0\linewidth]{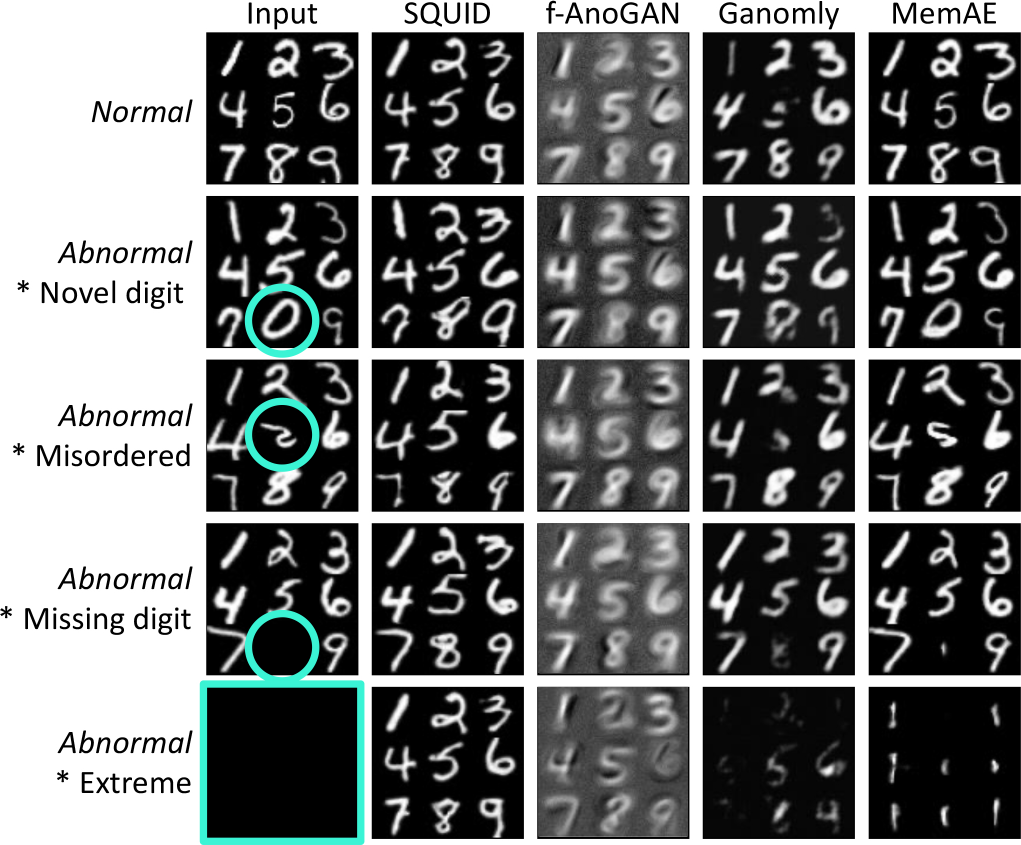}
    \caption{\textbf{Reconstruction results on \ourdataset.} 
    Our feature-level in-painting is more robust to amplified noise and pixel variance than the existing pixel-level in-painting methods.
    More visualization can be found in Appendix~\ref{sec:more_visual}.
    }
    \label{fig:toy_demo}
\end{figure}

\noindent\textbf{\textit{\ourdataset.}} We have created a synthetic dataset to verify our main idea, wherein the human anatomy is translated into Arabic digits one to nine in an in-grid placement (see examples in \figureautorefname~\ref{fig:introductory} and \figureautorefname~\ref{fig:toy_demo}).
The images containing digits in the correct order are considered ``normal''; otherwise, they are considered ``abnormal''. 
The types of simulated abnormalities include missing, misordered, flipped, and zero digit(s).
\ourdataset\ is particularly advantageous for radiography imaging for three reasons. 
\emph{First}, it simulates two unique properties of  radiography images, \ie spatial correlation and consistent shape.
\emph{Second}, annotating radiography images demands specialized expertise, but digits are easier for problem debugging. 
\emph{Third}, the ground truth of the simulated anomaly is readily accessible in \ourdataset, whereas it is hard to collect sufficient examples for each abnormal type in  radiography images.
The pseudocode for creating \ourdataset\ is in Appendix~\ref{sec:create_digitanatomy}.

\subsection{Public Benchmarks}

\noindent\textbf{\textit{ZhangLab Chest X-ray~\cite{kermany2018identifying}.}} 
This dataset contains healthy and pneumonia (as anomaly) images, \emph{officially} split into training and testing sets. 
The training set consists of 1,349 normal and 3,883 abnormal images;
the testing set has 234 normal and 390 abnormal images. 
We randomly separate 200 images (100 normal and 100 abnormal) from the training set as the validation set for hyper-parameter tuning. We resized all the images to $128\times128$.


\begin{table}[t]
\scriptsize
\centering
\caption{Benchmark results on the test sets of the two datasets.}
    \begin{tabular}{p{0.24\linewidth}p{0.14\linewidth}P{0.12\linewidth}P{0.12\linewidth}P{0.12\linewidth}}
        \emph{ZhangLab} & Ref~\&~Year & AUC~(\%) & Acc~(\%) & F1~(\%) \\
        \shline
        Auto-Encoder & - & 59.9 & 63.4 & 77.2  \\
        VAE~\cite{kingma2013auto} & Arxiv'13 & 61.8 & 64.0 & 77.4  \\
        Ganomaly~\cite{akcay2018ganomaly} & ACCV'18& 78.0 & 70.0 & 79.0  \\
        f-AnoGAN~\cite{schlegl2019f} & MIA'19 & 75.5 & 74.0 & 81.0 \\
        MemAE~\cite{gong2019memorizing}& ICCV'19 & 77.8$\pm$1.4 & 56.5$\pm$1.1 & 82.6$\pm$0.9 \\
        MNAD~\cite{park2020learning} & CVPR'20 & 77.3$\pm$0.9 & 73.6$\pm$0.7 & 79.3$\pm$1.1 \\
        SALAD~\cite{zhao2021anomaly} & TMI'21& 82.7$\pm$0.8 & 75.9$\pm$0.9 & 82.1$\pm$0.3  \\
        CutPaste~\cite{li2021cutpaste} & CVPR'21 &  73.6$\pm$3.9 & 64.0$\pm$6.5 & 72.3$\pm$8.9  \\
        PANDA~\cite{reiss2021panda} & CVPR'21 & 65.7$\pm$1.3 & 65.4$\pm$1.9 & 66.3$\pm$1.2  \\
        M-KD~\cite{salehi2020distillation} & CVPR'21 & 74.1$\pm$2.6 & 69.1$\pm$0.2 & 62.3$\pm$8.4  \\
        IF 2D~\cite{naval2021implicit} & MICCAI'21 & 81.0$\pm$2.8& 76.4$\pm$0.2& 82.2$\pm$2.7 \\
        PaDiM~\cite{defard2021padim} & ICPR'21 & 71.4$\pm$3.4 & 72.9$\pm$2.4 &80.7$\pm$1.2 \\
        IGD~\cite{chen2021deep} & AAAI'22 & 73.4$\pm$1.9 & 74.0$\pm$2.2 & 80.9$\pm$1.3 \\
        \hline
        \methodname  & - & \textbf{87.6$\pm$1.5} & \textbf{80.3$\pm$1.3} & \textbf{84.7$\pm$0.8}\\
    \end{tabular}
    \vfill
    \vspace{1em}
    \begin{tabular}{p{0.24\linewidth}p{0.14\linewidth}P{0.12\linewidth}P{0.12\linewidth}P{0.12\linewidth}}
        \emph{CheXpert} & Ref~\&~Year & AUC~(\%) & Acc~(\%) & F1~(\%) \\
        \shline
        Ganomaly~\cite{akcay2018ganomaly} & ACCV'18 & 68.9$\pm$1.4 & 65.7$\pm$0.2 & 65.1$\pm$1.9\\
        f-AnoGAN~\cite{schlegl2019f}& MIA'19 & 65.8$\pm$3.3 & 63.7$\pm$1.8 & 59.4$\pm$3.8 \\
        MemAE~\cite{gong2019memorizing}& ICCV'19 & 54.3$\pm$4.0 & 55.6$\pm$1.4 & 53.3$\pm$7.0 \\
        CutPaste~\cite{li2021cutpaste} & CVPR'21 & 65.5$\pm$2.2 & 62.7$\pm$2.0 & 60.3$\pm$4.6  \\
        PANDA~\cite{reiss2021panda} & CVPR'21 & 68.6$\pm$0.9 & 66.4$\pm$2.8 & 65.3$\pm$1.5  \\
        M-KD~\cite{salehi2020distillation} & CVPR'21 & 69.8$\pm$1.6 & 66.0$\pm$2.5 & 63.6$\pm$5.7  \\
        \hline
        \methodname  & - & \textbf{78.1$\pm$5.1} & \textbf{71.9$\pm$3.8} & \textbf{75.9$\pm$5.7}\\
    \end{tabular}
    
    \label{tab:chest_xray_benchmark}
\end{table}

\smallskip\noindent\textbf{\textit{Stanford CheXpert~\cite{irvin2019chexpert}.}} 
We conducted evaluations on the front-view PA images in the CheXpert  dataset, which account for a total of 12 different anomalies. 
In all front-view PA images, there are 5,249 normal and 23,671 abnormal images for training; 
250 normal and 250 abnormal images (with at least 10 images per disease type) from the training set were used for testing. We used the same hyper-parameters found in the ZhangLab experiments.



\subsection{Baselines and Metrics}
We considered a total number of \textbf{13} major baselines for direct comparison: Auto-Encoder, VAE~\cite{kingma2013auto}---the classic UAD methods; Ganomaly~\cite{akcay2018ganomaly}, f-AnoGAN~\cite{schlegl2019f}, IF~\cite{naval2021implicit},
SALAD~\cite{zhao2021anomaly}---the current state of the arts for medical imaging; and MemAE~\cite{gong2019memorizing}, CutPaste \cite{li2021cutpaste}, M-KD~\cite{salehi2021multiresolution}, PANDA~\cite{reiss2021panda}, PaDiM~\cite{defard2021padim}, IGD~\cite{chen2021deep}---the most recent UAD methods. We evaluated performance using standard metrics: Receiver Operating Characteristic (ROC) curve, Area Under the ROC Curve (AUC), Accuracy (Acc), and F1-score (F1). Unless explicitly specified, we trained all models from scratch for at least \emph{three} times independently.


\subsection{Implementation Details} 
We utilized common data augmentation strategies such as random translation within the $[-0.05, +0.05]$ range and a random scaling of $[0.95, 1.05]$. 
The Adam~\cite{kingma2014adam} optimizer was used with a batch size of 16 and a weight decay of $1e$-$5$. 
The learning rate was initially set to $1e$-$4$ for both the generator and the discriminator and then decayed to $2e$-$5$ in 1000 epochs following the cosine annealing scheduler. 
The discriminator is trained at every iteration, while the generator is trained every two iterations. 
We set loss weights as $\lambda_{\text{t}}=0.01$, $\lambda_{\text{s}}=10$, $\lambda_{\text{dist}}=0.001$, $\lambda_{\text{gen}}=0.005$, and $\lambda_{\text{dis}}=0.005$.
We divide the input images in $2\times2$ non-overlapping patches, fix the shortcut mask probability at $\rho=95\%$, and activate only the top $5$ similar patterns in the Gumbel Shrinkage. The impact of these hyper-parameters is studied in~\S\ref{sec:ablation}. 
The architectures of our generators and discriminator are detailed in Appendix~\ref{sec:detailed_architecture}.


\begin{figure}[t]
    \centering
    \includegraphics[width=1.\linewidth]{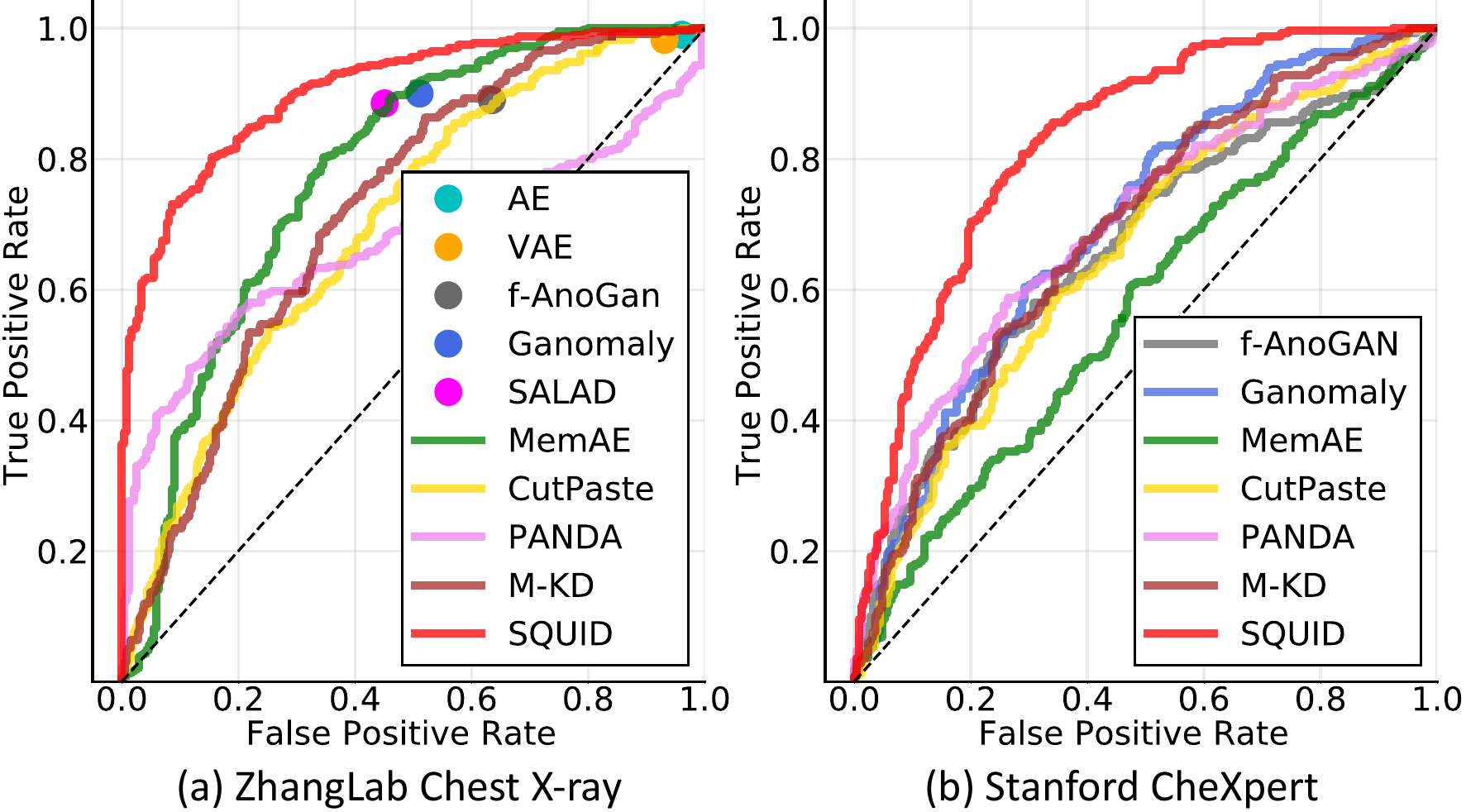}
    \caption{ROC curves comparison on the two datasets.}
    \label{fig:roc}
\end{figure}

\section{Results}
\label{sec:results}

\subsection{Interpreting \methodname\ on \ourdataset}
\label{sec:digit}

\figureautorefname~\ref{fig:toy_demo} presents qualitative results on \ourdataset\ to examine the capability of image reconstruction and to interpret the mistakes made by existing methods~\cite{schlegl2019f,akcay2018ganomaly,gong2019memorizing}. 
We deliberately inject anomalies (\eg novel, misordered, missing digits) into normal images (highlighted in light blue) and test if the model can reconstruct their normal counterparts. To raise the task difficulty, we also assess the reconstruction quality from a blank image (as an extreme case). 
In general, the images reconstructed by our \methodname\ carry more meaningful and indicative information than other baseline methods.
It is mainly attributed to our \emph{space-aware} memory, with which the resulting dictionary is associated with unique patterns as well as their spatial information.
Once an anomaly arises (\eg missing digit), the in-painting block will augment the abnormal feature to its normal counterpart by assembling top-$k$ most similar patterns from the dictionary.
Other methods, however, do not possess this ability, so they reconstruct defective images.
For instance, GAN-based methods (f-AnoGAN and Ganomly) tend to reconstruct an exemplar image averaged from the training examples.
MemAE performs relatively better due to its Memory Matrix, but it does not work well for the anomaly of missing digits and completely fails on the extreme anomaly attack.

\begin{figure}[t]
    \centering
    \includegraphics[width=1.\linewidth]{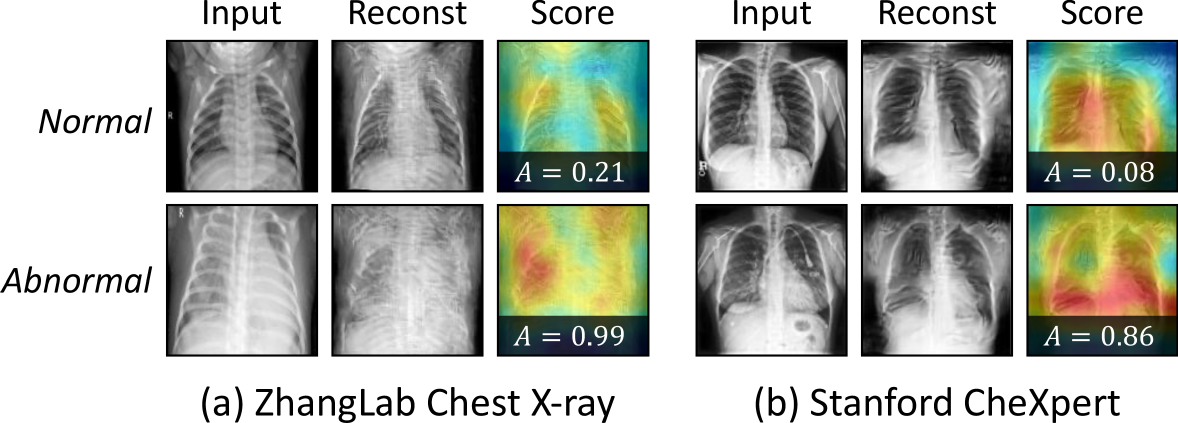}
    \caption{Reconstruction results of \methodname\ on the two datasets, associated with the corresponding anomaly scores (defined in \S\ref{sec:alert}).
    A larger score indicates a higher probability of being abnormal.
    More visualization can be found in Appendix~\ref{sec:more_visual}.
    }
    \label{fig:recon_results}
\end{figure}

\subsection{Benchmarking \methodname\ on Chest Radiography} \label{sec:benchmark_xray}

Our \methodname\ was mainly evaluated on two large-scale benchmarks: ZhangLab Chest X-ray and Stanford CheXpert and compared with a wide range of state-of-the-art counterparts. According to \tableautorefname~\ref{tab:chest_xray_benchmark}, \methodname\ achieves the most promising result on all metrics for both datasets. Specifically, \methodname\ outperforms the runner-up counterparts by at least $~5\%$ in AUC, $5\%$ in Accuracy. The highest F1 scores \methodname\ achieved, along with the ROC curves shown in \figureautorefname~\ref{fig:roc}, demonstrate that our method yields the best trade-off between sensitivity and specificity. Overall, the significant improvements observed with \methodname\ proved the effectiveness of our proposed techniques in this work.
In \figureautorefname~\ref{fig:recon_results}, we visualize the reconstruction results of \methodname\ on exemplary normal and abnormal images in the two datasets. For normal cases, \methodname\ can easily find a similar match in Memory Queue, achieving the reconstruction smoothly. For abnormal cases, the contradiction will arise by imposing forged normal patterns into the abnormal features. In this way, the generated images will vary significantly from the input, which will then be captured by the discriminator. We plot the heatmap of the discriminator (using Grad-CAM~\cite{selvaraju2017grad}) to indicate the most likely regions to appear anomalous. As a result, the reconstructed healthy images yield much lower anomaly scores than the diseased ones, validating the effectiveness of \methodname.

\smallskip\noindent\textbf{\textit{Limitation.}} We found \methodname\, in its current form, is not able to \emph{localize} anomalies at the pixel level precisely. It is understandable because our \methodname\ is an unsupervised method, requiring zero manual annotation for normal/abnormal images, unlike \cite{wang2017chestx,singh2017hide,zhang2018adversarial,tang2018attention,salehi2021multiresolution}. Those methods that compute pixel-level residuals for anomaly detection suffer from amplified noise in the input and reconstructed output.
Our in-painting strategy, however, is performed at the feature level and is more robust to pixel-level variance.

\begin{table}[t]
\scriptsize
\centering
\caption{
   Performance benefits from all the components in \methodname.} 
    \begin{tabular}{p{0.4\linewidth}P{0.13\linewidth}P{0.13\linewidth}P{0.13\linewidth}}
        Method & AUC~(\%) & Acc~(\%) & F1~(\%) \\
        \shline
        \emph{w/o} Space-aware Memory & 77.6$\pm$0.5& 75.5$\pm$0.5 & 82.5$\pm$0.6 \\
        \emph{w/o} In-painting Block & 80.9$\pm$2.1 & 75.8$\pm$1.5 & 81.6$\pm$1.3\\
        \emph{w/o} Gumbel Shrinkage & 81.1$\pm$0.9 & 77.6$\pm$0.9 & 81.3$\pm$0.8 \\
        \emph{w/o} Knowledge Distillation & 81.2$\pm$0.8 & 75.2$\pm$0.7 & 81.3$\pm$0.8 \\
        \emph{w/o} Stop Gradient & 81.7$\pm$4.3 & 76.7$\pm$2.8 & 82.5$\pm$1.6 \\
        \emph{w/o} Memory Queue & 82.5$\pm$1.1 & 78.6$\pm$0.9 & 81.7$\pm$1.1 \\
        \emph{w/o} Masked Shortcuts & 82.5$\pm$1.3 & 76.4$\pm$0.8 & 82.3$\pm$1.1 \\
        \emph{w/o} Decoder Memory & 82.9$\pm$1.2 & 77.4$\pm$1.1 & 81.2$\pm$0.5 \\
        \hline
        \textbf{Full \methodname} & \textbf{87.6$\pm$1.5} & \textbf{80.3$\pm$1.3} & \textbf{84.7$\pm$0.8}\\
    \end{tabular}
    \label{tab:component}
\end{table}

\begin{figure}[t]
    \centering
    \includegraphics[width=1.\linewidth]{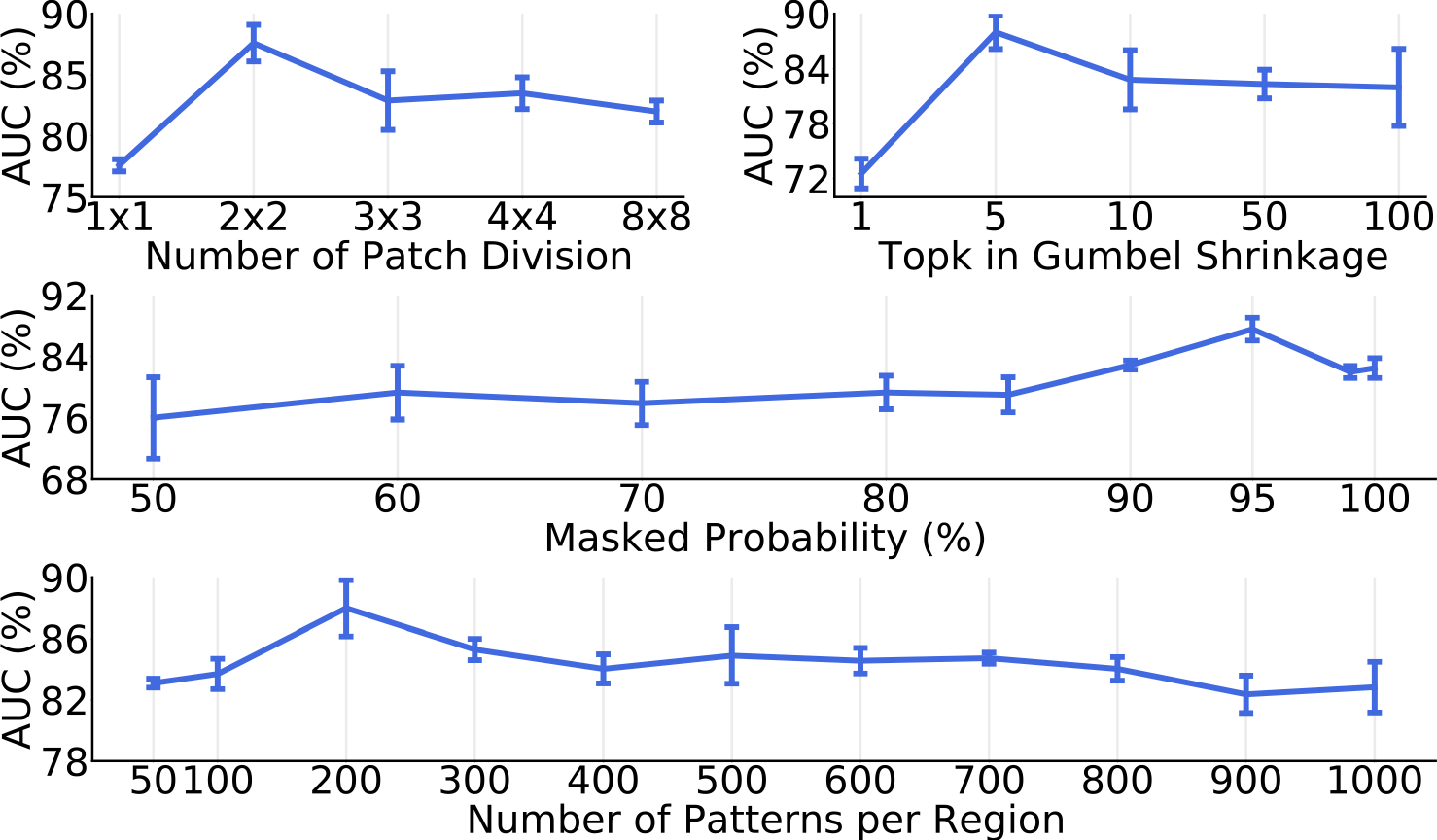}
    \caption{Hyper-parameter ablations.}
    \label{fig:ablation}
\end{figure}

\subsection{Ablating Key Properties in \methodname} 
\label{sec:ablation}

\noindent
\textbf{\textit{Component study.}} 
We examine the impact of components in \methodname\ by taking each one of them out of the entire framework. 
\tableautorefname~\ref{tab:component} shows that each component accounts for at least 5\% performance gain.
The space-aware memory ($+10.0$\%) and in-painting block ($+6.7$\%) are the top 2 most significant contributors, which underline our motivation and justification of the method development (\S\ref{sec:queue} and \S\ref{sec:inpaint}).
Although replacing Memory Queue with Memory Matrix could maintain a decent result (only dropped $5.1$\%), our Memory Queue presents a more trustworthy recovery of ``normal'' patterns in the image than Memory Matrix (MemAE~\cite{gong2019memorizing}), evidenced by~\figureautorefname~\ref{fig:toy_demo}.

\smallskip\noindent\textbf{\textit{Hyper-parameter robustness.}} After selecting the best hyper-parameters on the validation set, we here report the inference results on the testing set to study the robustness of different hyper-parameters in \figureautorefname~\ref{fig:ablation}. 
When input images are divided into a single patch, space-aware settings are not triggered, therefore yielding the worst performance. Although the spatial structures are relatively stable in most chest radiography, certain deviations can still be observed. Therefore, with small patches, object parts in one patch can easily appear in adjacent patches and be misdetected as anomalies. The number of topK activations in Gumbel softmax also impacts the performances. 
According to the AUC vs. the number of patterns in each Memory Queue region, we found that a small number of items is sufficient to support normal pattern querying in local regions. The best result is achieved by merely 200 items per region. When the item number exceeds 500 per region, AUC scores begin to drop continuously. AUC vs. the mask probability $\rho$ was further plotted to verify that enabling a limited number of feature skips ($\rho=95\%$) yields the best AUC score. The effectiveness of the in-painting will severely deteriorate if more features are allowed to be skipped ($\rho<90\%$).

\begin{figure}[t]
    \centering
    \includegraphics[width=\linewidth]{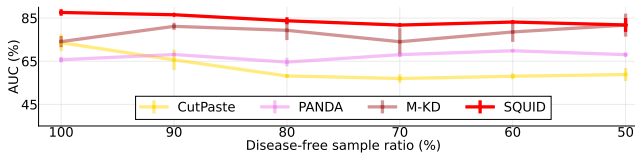}
    \caption{Results of mixing normal/abnormal training samples. 
    }
    \label{fig:posratio}
\end{figure}

\smallskip\noindent\textbf{\textit{Disease-free training requirement?}}
Unsupervised methods for medical anomaly detection are uncommon because the so-called UAD methods are \textit{not} ``unsupervised''---they must be trained on disease-free images only (\eg \cite{baur2021autoencoders}). In practice, cleaning up disease-free images relies on manual annotation (essentially, image-level healthy/diseased labels).
With disease-free sample ratio in the training set ranging from 100\% to 50\%, we have compared the robustness of \methodname\ with three competitive baselines (CutPaste \cite{li2021cutpaste}, PANDA \cite{reiss2021panda} and M-KD \cite{salehi2021multiresolution}).
\figureautorefname~\ref{fig:posratio} remarks that our proposed memory queue can tolerate the disease/healthy training ratio up to 50\% by automatically omitting minority anatomical patterns.
In contrast, CutPaste drops significantly as the percentage of normal images decreases; PANDA and M-KD can maintain the performance due to the use of pre-trained features. Interestingly, M-KD with mixed data even outperforms its vanilla training setting, although with considerable fluctuations. 

\section{Conclusion}

We present \methodname\ for unsupervised anomaly detection from  radiography images.
Qualitatively, we show that \methodname\ can taxonomize the ingrained anatomical structures into recurrent patterns; and in the inference, \methodname\ can identify anomalies accurately.
Quantitatively, \methodname\ is superior to predominant methods by over 5 points AUC on the ZhangLab dataset and 10 points AUC on the Stanford CheXpert dataset.
The outstanding results are attributable to our observation: \emph{Radiography imaging protocols focus on particular body regions, therefore producing images of great similarity and yielding recurrent anatomical structures across patients.} We synthesized the \ourdataset\ dataset to resemble key attributes of chest anatomy in  radiography images for prompting future method development.

\medskip\noindent\textbf{Acknowledgements.} This work was supported by the Lustgarten Foundation for Pancreatic Cancer Research and partially by the Patrick J. McGovern Foundation Award.
We appreciate the constructive suggestions from Yingda Xia, Yixiao Zhang, Jessica Han, Yingwei Li, Bowen Li, Huiyu Wang, Adam Kortylewski, and Sonomi Oyagi.

\clearpage
{\small
\bibliographystyle{ieee_fullname}
\bibliography{refs,zzhou}
}

\newpage
\clearpage
\appendix



\section{Architectures of \methodname}
\label{sec:detailed_architecture}

Our \methodname\ consists of an encoder, a student (main) generator, a teacher generator, and a discriminator. All of the network architectures are built with plain convolution, batch normalization, and ReLU activation layers only. The architecture details of the encoder are shown in~\tableautorefname~\ref{table:encoder_detail}. For an input  radiography image (sized of $128\times128$), we first divide it into $2\times2$ non-overlapping patches (sized of $64\times64$). The encoder then extracts the patch features.

As mentioned in~\S\ref{sec:framework}, the student and teacher generators were constructed identically. The only difference is that additional Memory Matrices are placed in the student generator. The architecture details of the student generator are shown in \tableautorefname~\ref{table:generator_detail}. Skip connections from the encoder are only enabled at such levels that Memory Matrices are used. After the last Memory Matrix, the non-overlapping patches are put back as a whole for further reconstruction.

As shown in \tableautorefname~\ref{table:discriminator_detail}, the discriminator was constructed in a more lightweight style. Note that the images are discriminated at their full resolution (\ie $128\times128$) rather than in patches.

\begin{table}[h]
\caption{Encoder structure in \methodname.}
\scriptsize
\centering
    \begin{tabular}{p{0.13\linewidth}P{0.13\linewidth}P{0.4\linewidth}}
        Level & \#Channels & Resolution  \\ 
        \shline
        Input & 1 & $(2\times2)\times(64\times64)$\\
        1 & 32 & $(2\times2)\times(32\times32)$\\
        2 & 64 & $(2\times2)\times(16\times16)$\\
        3 & 128 & $(2\times2)\times(8\times8)$\\
        4 & 256 & $(2\times2)\times(4\times4)$\\
    \end{tabular}
    
    \label{table:encoder_detail}
\end{table}

\begin{table}[h]
\caption{Student and teacher generator structures in \methodname. S\&M denotes the usage of skip connections and Memory Matrix.
    Note that there is no Memory Matrix placed in the teacher generator.
    }
\scriptsize
\centering
    \begin{tabular}{p{0.13\linewidth}P{0.13\linewidth}P{0.13\linewidth}P{0.3\linewidth}}
        Level & \#Channels & \emph{w/} S\&M &Resolution \\
        \shline
        4 & 256 & \checkmark &$(2\times2)\times(4\times4)$\\
        3 & 128 & \checkmark&$(2\times2)\times(8\times8)$\\
        2 & 64 & &$32\times32$\\
        1 & 32 & &$64\times64$\\
        Output & 1 & &$128\times128$\\
    \end{tabular}
    
    \label{table:generator_detail}
\end{table}

\begin{table}[h]
\caption{Discriminator structure in \methodname.}
\scriptsize
\centering
    \begin{tabular}{p{0.13\linewidth}P{0.13\linewidth}P{0.4\linewidth}}
        Level & \#Channels & Resolution  \\ 
        \shline
        Input & 1 & $128\times128$\\
        1 & 16 & $64\times64$\\
        2 & 32 & $32\times32$\\
        3 & 64 & $16\times16$\\
        4 & 128 & $8\times8$\\
        5 & 128 & $4\times4$\\
        Output & 1 & $1\times1$ \\ 
    \end{tabular}
    
    \label{table:discriminator_detail}
\end{table}

\section{Additional Results}
\label{sec:extensive_ablation_appendix}

\subsection{Extensive Ablation Studies}
In this section, we ablate three components in \methodname\ to fully validate their necessity and effectiveness.

\smallskip\noindent\textbf{(1) Convolutional~vs.~Transformer Layers:} In our proposed in-painting block, a transformer layer is used to aggregate the encoder extracted patch features, and the Memory Queue augmented ``normal'' features. However, one may wonder if a simple convolution layer can also suffice. We conducted experiments by replacing the transformer layer with a convolutional layer while preserving other structures. 

\smallskip\noindent\textbf{(2) Soft~vs.~Hard Masked Shortcuts:} In our proposed masked shortcut, skipped and in-painted features are aggregated using a binary gating mask. The intuitive question is whether such ``hard'' gating is necessary and a weighted ``soft'' addition can also achieve comparable results. To this end, instead of following Eq.~\ref{eq:masked_shortcut}, we conducted experiments by aggregating the patch features $\mathcal{F}$ through:
\begin{equation}
    \mathcal{F}' = (1-\rho)\cdot\mathcal{F} + \rho\cdot\texttt{inpaint}(\mathcal{F}),
\end{equation}
where $\rho$ was set to $95\%$, same as the best setting in \methodname.

\smallskip\noindent\textbf{(3) Pixel-level~vs.~Feature-level In-painting:} As discussed in~\S\ref{sec:inpaint}, raw images usually contain larger noise and artifacts than features, so we proposed to achieve the in-painting at the feature level rather than at the image level~\cite{li2020recurrent,pathak2016context,zhou2021models}. To validate our claim, we have conducted experiments on carrying out the in-painting at the pixel level. Instead of using a transformer layer to in-paint the extracted patch features, we randomly zeroed out parts of the input patches with 25\% probability and let \methodname\ in-paint the distorted input images. All other settings and objective functions remain unchanged. 

\smallskip\noindent\textbf{Summary:} The results of the above three additional ablative experiments are presented in \tableautorefname~\ref{tab:extensive_results}. Without using the transformer layer, masked shortcut, and feature-level in-painting as proposed, the AUC, Acc, and F1 scores decreased by at least 8\%, 4\%, and 3\%, respectively, compared with the full \methodname\ setting.


\begin{table}[t]
\caption{
    The extensive results indicate that all proposed techniques in \methodname\ are essential for a high overall performance.} 
\scriptsize
\centering
     \begin{tabular}{p{0.4\linewidth}P{0.13\linewidth}P{0.13\linewidth}P{0.13\linewidth}}
        Method & AUC~(\%) & Acc~(\%) & F1~(\%) \\
        \shline
        Convolution Layers & 76.9$\pm$3.3 & 74.2$\pm$3.3 & 80.7$\pm$2.7 \\
        Transformer Layers ($\Delta$) & \textcolor{Highlight}{$\uparrow$10.7} & \textcolor{Highlight}{$\uparrow$6.1} & \textcolor{Highlight}{$\uparrow$4.0} \\
        \hline
        Soft Masked Shortcut  & 79.7$\pm$3.4& 76.1$\pm$2.7 & 80.7$\pm$2.3 \\
        Hard Masked Shortcut ($\Delta$) & \textcolor{Highlight}{$\uparrow$7.9} & \textcolor{Highlight}{$\uparrow$4.2} & \textcolor{Highlight}{$\uparrow$4.0}\\
        \hline
        Pixel-level In-painting & 79.1$\pm$0.4 & 74.4$\pm$1.6 & 81.3$\pm$0.9 \\
        Feature-level In-painting ($\Delta$) & \textcolor{Highlight}{$\uparrow$8.5} & \textcolor{Highlight}{$\uparrow$5.9} & \textcolor{Highlight}{$\uparrow$3.4} \\
        \hline
        \textbf{Full \methodname} & \textbf{87.6$\pm$1.5} & \textbf{80.3$\pm$1.3} & \textbf{84.7$\pm$0.8}\\
    \end{tabular}

    \label{tab:extensive_results}
\end{table}

\subsection{Patch-MemAE}
MemAE~\cite{gong2019memorizing} with Memory Matrix is the primary baseline that we considered in this work. To further verify the effectiveness of our proposed space-aware setting, we trained additional MemAE models on patches segmented from different spatial location of input images. These multiple space-specific models were trained separately with their unique space-specific patches and were then evaluated through an ensemble style to compare with our \methodname. The results are reported in Table~\ref{tab:patch_memae}.

The results of the this experiment indicate that although improvements can be observed on AUC and Acc, such space-specific ensemble upgrade still performs inferior than \methodname. Moreover, we found such ensemble of models demands a much \textit{higher} degree of computational costs (\emph{4$\times$} more than ours), while in our work, we proposed to encode this spatial information into the feature dictionary, ultimately requiring only one model. Both effectiveness and efficiency are pronounced.

\begin{table}[t]
\caption{
    We apply space-specific strategy to one of the strongest counterparts (MemAE~\cite{gong2019memorizing}).
    In addition, the ensemble of spatial-aware models demands a \textit{higher} degree of computational costs (4$\times$ more than ours), while our work proposed to encode this spatial information into the feature dictionary, ultimately requiring only one model---its efficiency is pronounced.} 
\scriptsize
\centering
     \begin{tabular}{p{0.4\linewidth}P{0.13\linewidth}P{0.13\linewidth}P{0.13\linewidth}}
        Method & AUC~(\%) & Acc~(\%) & F1~(\%) \\
        \shline
        MemAE~\cite{gong2019memorizing}& 77.8$\pm$1.4 & 56.5$\pm$1.1 & 82.6$\pm$0.9  \\
        Patch-MemAE ($\Delta$) &  \textcolor{Highlight}{$\uparrow$0.5} & \textcolor{Highlight}{$\uparrow$18.5} & \textcolor{red}{$\downarrow$1.3} \\
        \hline
        \textbf{Full \methodname} & \textbf{87.6$\pm$1.5} & \textbf{80.3$\pm$1.3} & \textbf{84.7$\pm$0.8}\\
    \end{tabular}
    
    \label{tab:patch_memae}
\end{table}

\section{Creating \ourdataset}
\label{sec:create_digitanatomy}

The pseudocode of creating our new benchmark dataset (\ourdataset\ in \S\ref{sec:new_benchmark}) is provided in Algorithm~\ref{alg:digitanatomy}. In practice, we have implemented the algorithm into an off-the-shelf data loader that can be amended to many other different datasets (\eg SVHN, CIFAR, ImageNet).

\begin{algorithm}[t]
\footnotesize
   \caption{Creating \ourdataset}
   \label{alg:digitanatomy}
    \definecolor{codeblue}{rgb}{0.25,0.5,0.5}
    \lstset{
      basicstyle=\fontsize{8.2pt}{8.2pt}\ttfamily\bfseries,
      commentstyle=\fontsize{8.2pt}{8.2pt}\color{codeblue},
      keywordstyle=\fontsize{8.2pt}{8.2pt},
    }
\begin{lstlisting}[language=python]
# a function to pick random digit instances
def pick_random(class, single_digits):
    # random pick an image with size: [28, 28]
    pick_digit = random.choice(single_digits[class])
    return pick_digit


# load MNIST digits with shape: [10, 1000, 28, 28]
single_digits = load_MNIST() 

# all possible conditions
conditions = ['normal', 'missing',\
              'misorder', 'flipped', 'novel']

output = torch.zeros(3, 28, 3, 28)

# loop over digit 1-9 in order
for idx in range(1,10):

  # randomly pick a condition
  condition = random.choice(conditions)
  
  if condition == 'normal':
    digit = pick_random(idx, single_digits)
  # anatomy of missing digit
  elif condition == 'missing':
    digit = torch.zeros(28,28)
  # anatomy of disorder digit
  elif condition == 'misorder':
    ridx = randrom.randint(1,10)
    digit = pick_random(ridx, single_digits)
  # anatomy of flipped digit
  elif condition == 'flipped':
    digit = pick_random(idx, single_digits)
    digit = digit[::-1,::-1]
  # anatomy of novel digit
  elif condition == 'novel':
    digit = pick_random(0, single_digits)
    
  output[idx // 3, :, idx % 3, :] = digit

# combine all patches together
output = output.view(28 * 3, 28 * 3)
\end{lstlisting}

\end{algorithm}

\section{Visualization Results}
\label{sec:more_visual}

\subsection{Visualizations on \ourdataset}

\begin{figure*}
    \centering
    \includegraphics[width=1.0\linewidth]{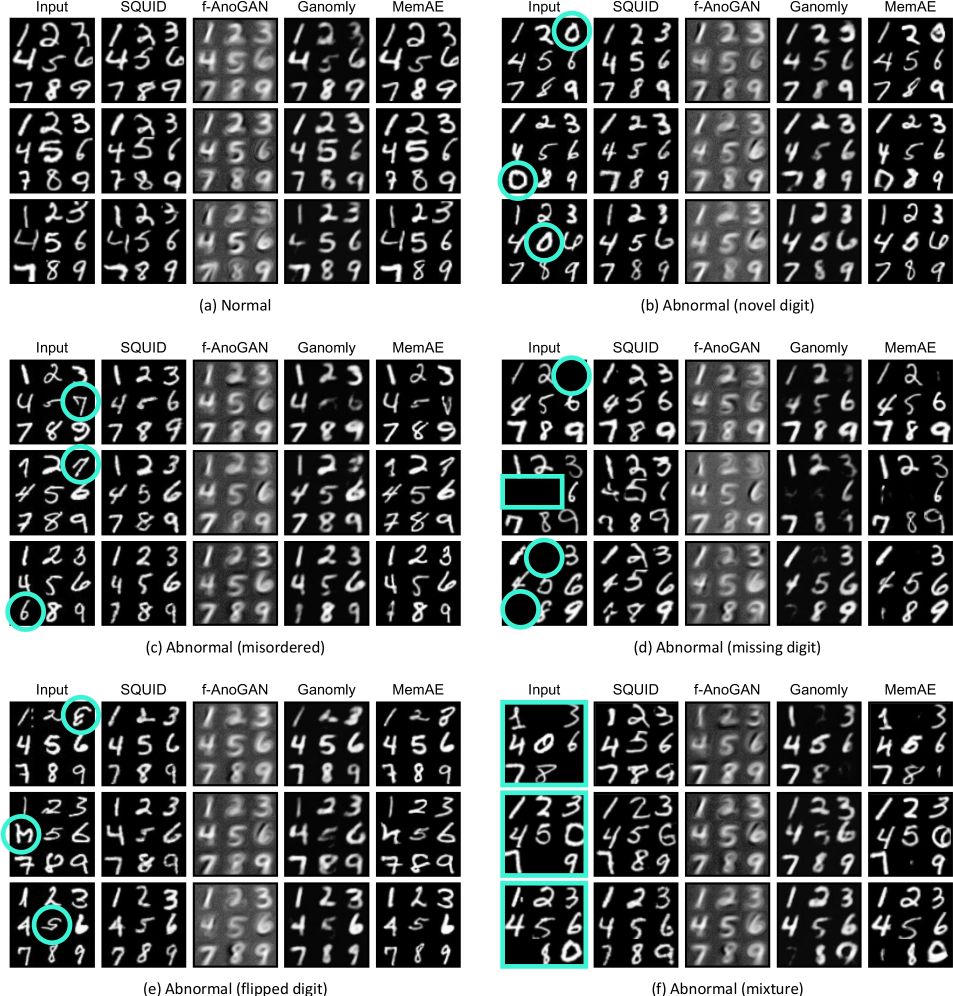}
    \caption{Comparisons of reconstruction results on \ourdataset\ of our \methodname, f-AnoGAN~\cite{schlegl2019f}, Ganomaly~\cite{akcay2018ganomaly}, and MemAE~\cite{gong2019memorizing}. Anomalies are highlighted in light blue.
    }
    \label{fig:more_interpretation}
\end{figure*}

More reconstruction results of \methodname\ and the compared methods~\cite{akcay2018ganomaly,schlegl2019f,gong2019memorizing} are shown in \figureautorefname~\ref{fig:more_interpretation}. Our observations from these additional results are aligned with the ones discussed in \S\ref{sec:digit}. \methodname\ can capture \emph{every} appearing anomaly (highlighted in light blue) in the images and augment them back to the normal closest forms. On the contrary, although MemAE restores the normal digits the best, it is limited in detecting a few anomaly types (\eg misordered and missing digits). Ganomaly is not able to perfectly recover the normal digits and also cannot generate meaningful reconstructions on the abnormal ones. f-AnoGAN, on the other hand, memorizes and generates an exemplary normal pattern that fails to respond to different inputs. 

\subsection{Visualizations on Chest Radiography}

\begin{figure*}
    \centering
    \includegraphics[width=1.0\linewidth]{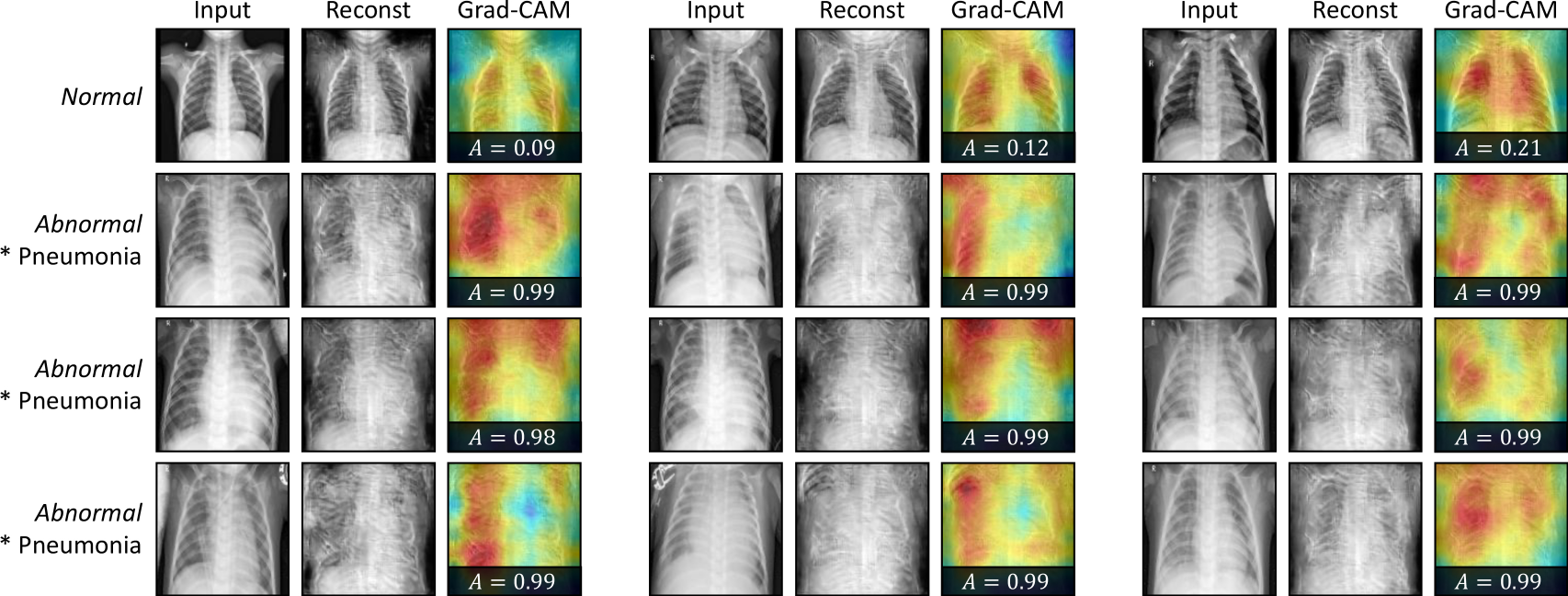}
    \caption{Reconstruction results of \methodname\ on the ZhangLab Chest X-ray dataset. The corresponding Grad-CAM heatmaps along with anomaly scores are shown as well. 
    }
    \label{fig:more_anomaly_score_zhanglab}
\end{figure*}

\begin{figure*}
    \centering
    \includegraphics[width=1.0\linewidth]{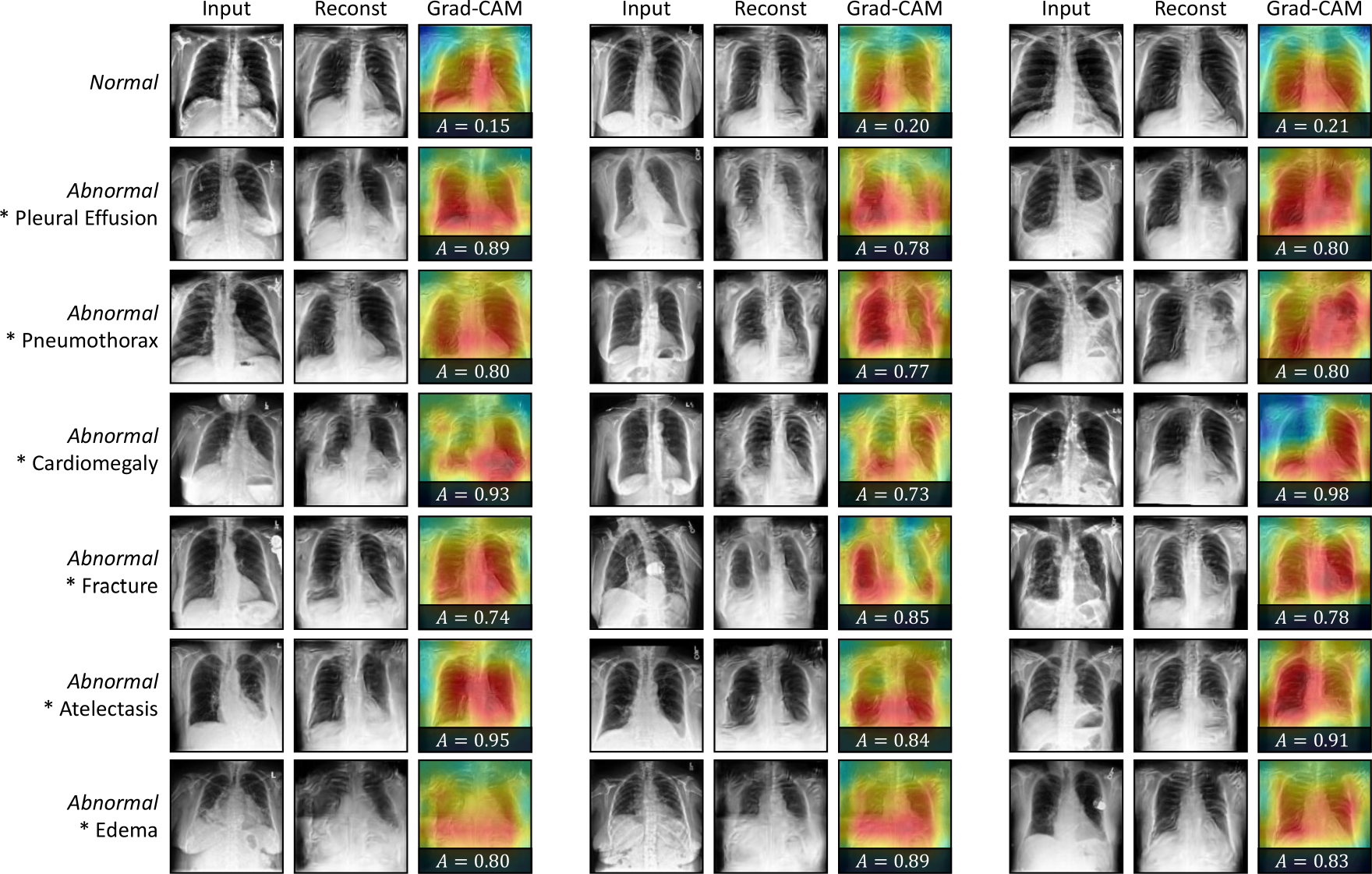}
    \caption{Reconstruction results of \methodname\ on the Stanford CheXpert dataset. Different disease types are separated into different rows. The corresponding Grad-CAM heatmaps along with anomaly scores are shown as well. 
    }
    \label{fig:more_anomaly_score_chexpert}
\end{figure*}

\figureautorefname~\ref{fig:more_anomaly_score_zhanglab} and \figureautorefname~\ref{fig:more_anomaly_score_chexpert} show more reconstruction results of our \methodname\ on the ZhangLab Chest X-ray and Stanford CheXpert datasets. We observed that our method is capable of translating the input image to its ``normal'' counterpart and assigning larger anomaly scores to abnormal cases.

When inputting normal images, \methodname\ will try to reconstruct the inputs as well as possible. Due to the usage of memory modules, our framework could hardly degenerate to function as an identity mapping from inputs to outputs. Therefore, the reconstruction of normal inputs cannot perfectly recover every single detail. 

When inputting abnormal images, \methodname\ will make larger impacts by combining previously seen normal features together into such abnormal ones. Since the generator is not trained on such hybrid features, the reconstruction results could demonstrate more obvious artifacts and blurs.  

After our framework converges, the optimized discriminator can perceptually capture such inconsistencies between reconstructed normal and abnormal images and achieve anomaly detection.

\end{document}